\documentclass[10pt,twocolumn,letterpaper]{article}

\usepackage{cvpr}              

\usepackage{graphicx}
\usepackage{amsmath}
\usepackage{amssymb}
\usepackage{booktabs}

\usepackage[pagebackref,breaklinks,colorlinks]{hyperref}

\usepackage[capitalize]{cleveref}
\crefname{section}{Sec.}{Secs.}
\Crefname{section}{Section}{Sections}
\Crefname{table}{Table}{Tables}
\crefname{table}{Tab.}{Tabs.}

\def\cvprPaperID{3483} 
\def\confName{CVPR}
\def\confYear{2023}

\usepackage{color}
\usepackage{xcolor}
\definecolor{deepGreen}{RGB}{0,153,0}
\definecolor{orange}{RGB}{255,125,0}
\def\red#1{\textcolor[rgb]{1,0,0}{#1}}

\newcommand{\keypoint}[1]{\vspace{0.1cm}\noindent\textbf{#1}\;}
\newcommand{\cut}[1]{}
\usepackage{multirow}
\usepackage{amsmath}
\usepackage{amssymb}
\usepackage{mathtools}
\usepackage{arydshln}
\usepackage{soul}
\usepackage{nicefrac}
\usepackage{booktabs}
\usepackage{stmaryrd}
\usepackage{caption}
\usepackage{graphicx}
\usepackage{colortbl}
\definecolor{Gray}{gray}{0.9}
\usepackage[accsupp]{axessibility}

\begin{document}

\title{Sketch2Saliency: Learning to Detect Salient Objects from Human Drawings \vspace{-0.6cm}}

\author{Ayan Kumar Bhunia\textsuperscript{1} \hspace{.5cm}  Subhadeep Koley\textsuperscript{1,2} \hspace{.5cm} Amandeep Kumar\thanks{Interned with SketchX} \hspace{.5cm} Aneeshan Sain\textsuperscript{1,2} \\ \hspace{.3cm}  Pinaki Nath Chowdhury\textsuperscript{1,2} \hspace{.4cm}
Tao Xiang\textsuperscript{1,2}\hspace{.4cm}  Yi-Zhe Song\textsuperscript{1,2} \\
\textsuperscript{1}SketchX, CVSSP, University of Surrey, United Kingdom.  \\
\textsuperscript{2}iFlyTek-Surrey Joint Research Centre on Artificial Intelligence.\\
{\tt\small \{a.bhunia, s.koley, a.sain, p.chowdhury, t.xiang, y.song\}@surrey.ac.uk}
\vspace{-0.6cm}
}
\maketitle

\begin{abstract}
\vspace{-0.2cm}
Human sketch has already proved its worth in various visual understanding tasks (e.g., retrieval, segmentation, image-captioning, etc). In this paper, we reveal a new trait of sketches -- that they are also salient. This is intuitive as sketching is a natural attentive process at its core. More specifically, we aim to study how sketches can be used as a weak label to detect salient objects present in an image. To this end, we propose a novel method that emphasises on how “salient object” could be explained by hand-drawn sketches. To accomplish this, we introduce a photo-to-sketch generation model that aims to generate sequential sketch coordinates corresponding to a given visual photo through a 2D attention mechanism. Attention maps accumulated across the time steps give rise to salient regions in the process. Extensive quantitative and qualitative experiments prove our hypothesis and delineate how our sketch-based saliency detection model gives a competitive performance compared to the state-of-the-art.
\end{abstract}


\vspace{-0.6cm}
\section{Introduction}
\vspace{-0.1cm}
As any reasonable drawing lesson would have taught you -- sketching is an \emph{attentive} process \cite{hertzmann2020line}. This paper sets out to prove just that but in the context of computer vision. In particular, we show that\cut{the} attention information\cut{that is} inherently embedded in sketches can be cultivated to learn image saliency \cite{jiang2021saliency, wang2017learning, zhang2018progressive, zeng2019multi}.

Sketch research has flourished in the past decade, particularly with the proliferation of touchscreen devices. Much of the utilisation of sketch has anchored around its human-centric nature, in that it naturally carries personal style \cite{sain2021stylemeup, bhunia2022adaptive}, subjective abstraction \cite{bhunia2020sketch, sain2020cross}, human creativity \cite{ge2020creative},\cut{just} to name a few. Here\cut{In this paper,} we study a sketch-specific trait that has been ignored to date -- \cut{that}sketch is also salient.

The human visual system has evolved over millions of years to develop the ability to attend \cite{yu2016sketch, ha2017neural}. This attentive process is ubiquitously reflected in language (i.e., how we describe visual concepts) and art (i.e., how artists attend to different visual aspects). The vision community has also invested significant effort to model this attentive process, in the form of saliency detection \cite{yang2013saliency, zeng2019multi, yu2021structure, piao2021critical, wang2021salient}. The paradox facing the saliency community is however that the attention information has never been present in photos to start with -- photos are mere collections of static pixels. It then comes with no surprise that most prior research has resorted to a large amount of pixel-level annotation.

\begin{figure}[!t]
\begin{center}
  \includegraphics[width=1\linewidth]{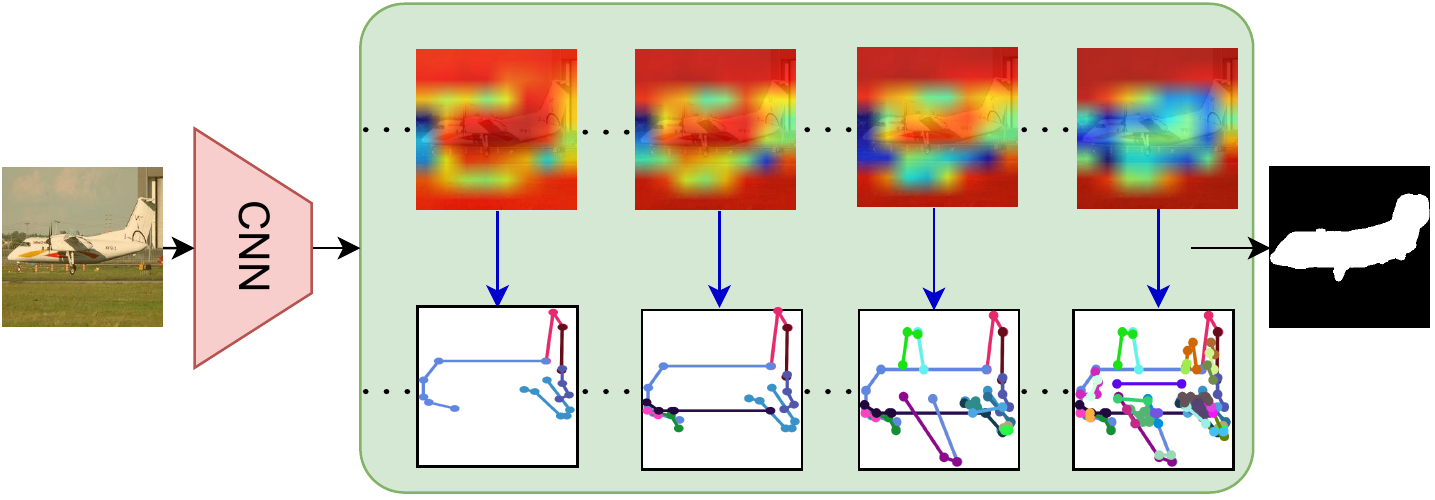}
\end{center}
\vspace{-0.7cm}
 \caption{Sequential \emph{photo-to-sketch} generation with 2D-attention to leverage sketch as a weak label for salient object detection. Aggregated 2D attention-maps till a particular instant are shown.}
  \vspace{-0.5cm}
\label{fig:Self_fig1.pdf}
\end{figure}

Although fully-supervised frameworks \cite{li2015visual, liu2016dhsnet, zhang2018progressive, chen2018reverse} have been shown to produce near-perfect saliency maps, its widespread adoption is largely bottlenecked by this \cut{requirement}need for annotation. To deal with this issue, a plethora of semi/weakly-supervised methods have been introduced, which attempt to use captions \cite{zeng2019multi}, class-labels \cite{oh2017exploiting}, foreground mask \cite{wang2017learning}, class activation map (CAM) \cite{wei2017object}, bounding boxes \cite{liu2021weakly}, scribbles \cite{zhang2020weakly} as weak labels. We follow this push to utilise labels, but importantly introduce sketch to the mix, and show it is a \cut{better/}competitive label modality because of the inherently embedded attentive information it possesses.

\cut{However, utilisation of sketch as weak label for saliency detection is non-trivial. Sketch, primarily being sequential in nature, portrays significant modality gap with photos. Also, the sparse and abstract nature of hand-drawn sketches must be dealt gracefully to extract significant visual intelligence insight from it. In particular, sketch is not an edge-wise tracing of photos. Therefore, classical template matching algorithms can not be applied in our purpose. Therefore, we are after a framework that can connect the sketch with the input photo space through which we can obtain pixel-wise importance value depicting the saliency map.}

Utilising sketch as a weak label for saliency detection is nonetheless non-trivial. Sketch, primarily being abstract and sequential \cite{ha2017neural} in nature, portrays significant modality gap with photos. Therefore, we seek to build a framework that can connect sketch and photo domains via some auxiliary task. For that, we take inspiration from an actual artistic sketching process, where artists \cite{zheng2018strokenet} attend to certain regions on an object, then render down the strokes on paper. We thus propose \emph{photo-to-sketch generation}, where given a photo we aim\cut{the aim is} to generate a sketch\cut{in a} stroke-by-stroke\cut{manner}, as an auxiliary task to bridge the two domains.

However effective in bridging the domain gap, this generation process by default does not generate pixel-wise importance values depicting a saliency map. To circumvent this problem, we make clever use of a cross-modal 2D attention module inside the sketch decoding process, which naturally predicts a local saliency map at each stroke -- exactly akin to how artists refer back to the object before rendering the next stroke.

More specifically, the proposed photo-to-sketch generator is an encoder-decoder model that takes an RGB photo as input and produces a sequential sketch. The model is augmented with a 2D attention mechanism that importantly allows the model to focus on visually salient regions of the photo associated with each stroke during sketch generation. In doing so, the attention maps accumulated over the time steps of sequential sketch generation would indicate the regions on the photo that were of utmost importance. See Fig.~\ref{fig:Self_fig1.pdf} for an illustration\cut{of this process}. To further address the domain gap in supervision between pixel-wise annotation and sketch labels, we propose an additional equivariance loss to gain robustness towards perspective deformations \cite{hung2019scops} thus improving overall performance.

In our experiments we firstly report the performance of saliency maps directly predicted by the network, without convolving it with any ad hoc post-processing that is commonplace in the literature \cite{zeng2019multi}. This is to spell out the true effects of using sketch for saliency detection, which is our main contribution. To further evaluate its competitiveness against other state-of-the-arts though, we also plug in our photo-to-sketch decoder in place of the image-captioning branch of a multi-source weak supervision framework that uses class-labels and text-description for saliency learning \cite{zeng2019multi}. We train our network on the Sketchy dataset \cite{sangkloy2016sketchy} consisting of photo-sketch pairs. It is worth noting that the training data was not intentionally collected with saliency detection in mind, rather the annotators were asked to draw a sketch that depicts the photo shown to them. This again strengthens our argument that sketch implicitly encodes saliency information which encouraged us to use it as a weak label at the first place.

In summary, our major contributions are: (a) We for the first time demonstrate the success of using sketch as a weak label in salient object detection. (b) To this end, we make clever use of sequential photo-to-sketch generation framework involving an auto-regressive decoder with 2D attention for saliency detection.  (c) Comprehensive quantitative and ablative experiments delineate that our method offers significant performance gain over weakly-supervised state-of-the-arts. Moreover, this is the first work in vision community that validates the intuitive idea of \emph{``Sketch is Salient"} through a simple yet effective framework.

\vspace{-0.2cm}
\section{Related Works}
\vspace{-0.1cm}
\noindent \textbf{Sketch for Visual Understanding:}
Hand-drawn sketches, which inherit the cognitive potential of human intelligence \cite{hertzmann2020line},
have been used in image retrieval \cite{bhunia2021more, bhunia2020sketch, yu2016sketch, PartialSBIR, sain2023exploiting, sain2023clip}, generation \cite{koley2023picture, ghosh2019interactive}, editing \cite{yang2020deep}, segmentation \cite{hu2020sketch},  image-inpainting \cite{xie2021exploiting}, video synthesis \cite{li2021deep}, representation learning \cite{wang2021sketchembednet}, 3D shape modelling \cite{zhang2021sketch2model}, augmented reality \cite{yan2020interactive}, among others \cite{xu2020deep, chowdhury2023scenetrilogy}. Due to its commercial importance, sketch-based image retrieval (SBIR) \cite{dutta2019semantically, sain2021stylemeup, collomosse2019livesketch, yelamarthi2018zero, shen2018zero,Sketch3T} has seen considerable attention within sketch-related research where the objective is to learn a sketch-photo joint-embedding space through various deep networks \cite{bhunia2020sketch, collomosse2019livesketch, sampaio2020sketchformer} for retrieving image(s) based on a sketch query.
In recent times, sketches have been employed to design aesthetic pictionary-style gaming \cite{sketchxpixelor} and to mimic the creative angle \cite{ge2020creative} of human understanding.
Growing research on the application of sketches to visual interpretation tasks establishes the representative \cite{wang2021sketchembednet} as well as discriminative \cite{DoodleIncremental, chowdhury2023what} ability of sketches to characterise the corresponding visual photo. Set upon this hypothesis, we aim to explore how saliency \cite{li2015visual} could be explained by sparse sketches \cite{sketch2vec}.


\vspace{0.05cm}
\noindent \textbf{Salient Object Detection:} Being pivotal to various computer vision tasks, saliency detection \cite{wang2021salient} has gained serious research attention \cite{yang2013saliency, zeng2019multi} as a pre-processing step in various downstream tasks.
%
In general, recent deep learning based saliency detection frameworks can be broadly categorised into – (i) fully supervised methods \cite{li2015visual, liu2016dhsnet, zhang2018progressive, chen2018reverse}, and (ii) weakly/semi-supervised methods \cite{wei2017object, liu2021weakly, zeng2019multi, yu2021structure}.
While most fully-supervised methods make use of pixel-level semantic labels for training, different follow-up works improved the performance by modelling multi-scale features with super-pixels \cite{li2015visual}, deep hierarchical network \cite{liu2016dhsnet} with progressive refinement, attention-guided network \cite{zhang2018progressive, chen2018reverse}.
As fully-supervised performance comes at the high cost of pixel-level labelling, various weakly-supervised methods were presented which utilise caption/tags \cite{zeng2019multi, wang2017learning}, bounding boxes \cite{liu2021weakly}, scribbles \cite{zhang2020weakly}, centroid detection \cite{tian2020weakly}, CAMs \cite{wei2017object} and foreground region masks \cite{wang2017learning} as weak labels \cite{zeng2019joint}. However, unlike class-agnostic pixel-level saliency labels, image feature or category label supervisions often lack reliability \cite{zeng2019multi}. Moreover, there exists added computational overhead \cite{zeng2019multi, zhang2020weakly} while calculating such weak labels. In this work, we take a completely different idiosyncratic approach of using amateur sketches \cite{sangkloy2016sketchy} as a means of weak supervision for learning image level saliency. Opposed to cumbersome architectures \cite{zeng2019multi, zhang2020weakly}, our simple framework validates and puts forth the potential of sketch as weak labels for saliency detection.

\noindent \textbf{Photo to Sketch Generation:}
Given our hypothesis of using sketch as a label in the output space, the genre of literature that connects image (photo) in the input space with sketch in the output space is photo-to-sketch generation \cite{liu2020unsupervised, song2018learning, bhunia2021more, cao2019ai} process.
While treating photo-to-sketch generation as image-to-image translation \cite{liu2020unsupervised}
lacks the capability of modelling the stroke-by-stroke nature of human drawing, we focus on  image-to-sequence generation pipeline, similar to Image Captioning \cite{xu2015show} or Text Recognition \cite{bhunia2021joint}, which involves a convolutional encoder followed by a RNN/LSTM decoder. Taking inspiration from the seminal work of Sketch-RNN \cite{ha2017neural}, several sequential photo-to-sketch generation pipelines \cite{song2018learning, bhunia2021more, cao2019ai} have surfaced in the literature in recent times, of which  \cite{bhunia2021more} made use of photo-to-sketch generation for semi-supervised fine-grained SBIR. However, these  prior works do not answer the question of how sequential sketch vectors could be leveraged for saliency detection, where the objective is to predict pixel-wise scalar importance value on the input image space.



\noindent \textbf{Attention for Visual Understanding:}
Following the great success in language translation \cite{luong2015effective} under natural language processing (NLP), attention mechanism has enjoyed recognition in vision related tasks as well. \emph{Attention} as a mechanism for reallocating the information according to the significance of the activation \cite{xu2015show}, plays a pivotal role in the modelling the human visual system-like behaviours. Besides incorporating the attention mechanism inside CNN architectures \cite{hu2018squeeze, zhang2019self}, multi-head attention is now an essential building block of the transformer \cite{vaswani2017attention} architecture which is rapidly growing in both NLP and computer vision literature.  Our attention scheme is more inclined to the design of image captioning \cite{xu2015show} works. Moreover, we employ attention mechanism to model the saliency map inside a cross-modal photo-to-sketch generation architecture.



\vspace{-0.2cm}
\section{Sketch for Salient Object Detection}
\label{sec:method}
\begin{figure}[!hbt]
\includegraphics[width=\columnwidth]{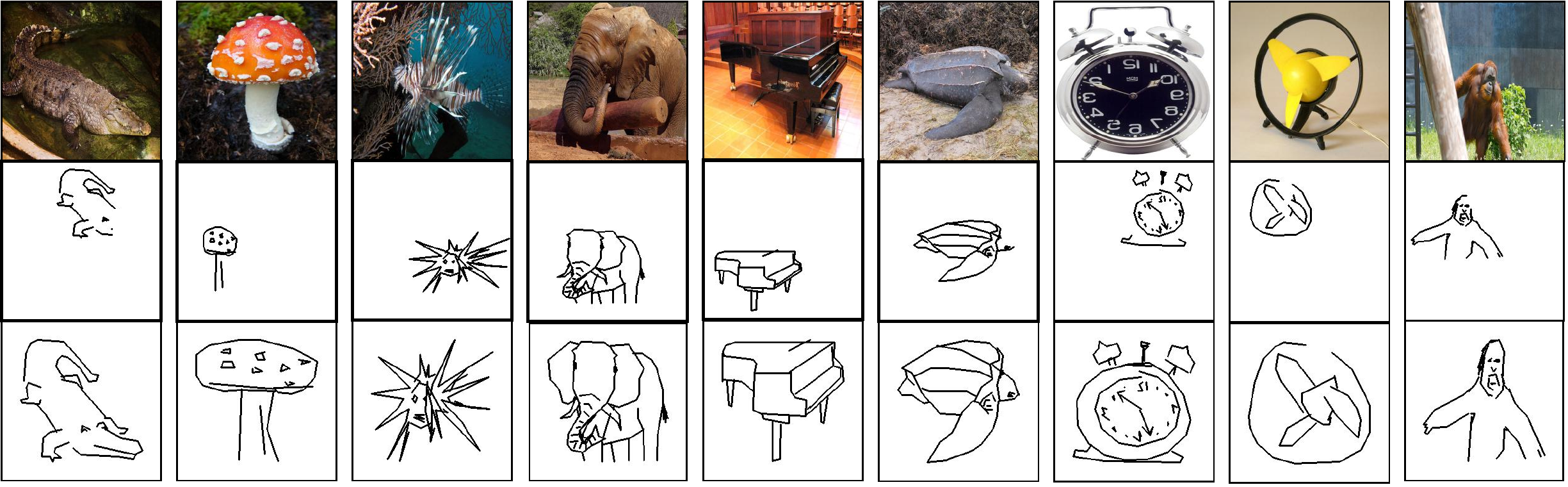}
\vspace{-0.8cm}
\caption{Example photo-sketch pairs with original, absolute coordinate rasterised and scale normalised photos being at the top, middle and bottom respectively. It is evident that photo and sketches are not aligned even after scale normalisation, as sketch is not a pixel-wise tracing of photo edge map. This invalidates the use of classical template matching algorithms \cite{brunelli2009template} for our problem.
}
\vspace{-0.4cm}
\label{example_fig}
\end{figure}

\vspace{-0.2cm}
\keypoint{Overview:} To learn the saliency from sketch labels, we have access to photo-sketch pairs as $\mathcal{D} = \{(P_i, S_i)\}_{i}^{N}$, where every sketch corresponds to its paired photo. A few examples are shown in Fig.~\ref{example_fig}.  Unlike a photo $P$ of size $\mathbb{R}^{H \times W \times 3}$, sketch holds dual-modality \cite{sketch2vec} of representation -- raster and vector sequence. While in raster modality sketch can {be} represented as a spatial image (like a photo), the same sketch can be depicted as a sequence of coordinate points in vector modality. As raster modality lacks the  stroke-by-stroke sequential hierarchical \cite{ha2017neural} information (pixels are not segmented into strokes), we use a sketch vector consisting of pen states $S = \{v_1, v_2, \cdots v_T\}$ where $T$ is the length of the sequence. Here, every point consists of a 2D absolute coordinate in a $H \times W$ canvas as $v_i = (x_i, y_i, b_i)$, where $b_i \in \{0, 1\}$ represents the start of a new stroke, i.e. stroke token.

In salient object detection \cite{li2015visual}, the objective is to obtain a saliency map $S_M$ of size $\mathbb{R}^{H \times W}$, where every scalar value  $S_M(i,j)$ represents the importance value of that pixel belonging to the salient image region. Therefore, the research question arises as to how can we design a framework that shall learn to predict a saliency map $S_M$ representing the important regions of an input image, using cross-modal training data (i.e., sketch vector-photo pairs). Towards solving this, we design an image-to-sequence generation architecture consisting of a 2D attention module between a convolutional encoder and a sequential decoder.

\vspace{-0.2cm}

\vspace{-0.2cm}
\subsection{Model Architecture}
\vspace{-0.1cm}
The intuition behind our architecture is that while generating every sketch coordinate at a particular time step, the attentional decoder should \emph{``look back''} at the relevant region of the input photo via an \emph{attention-map}, which when gathered across the whole output sketch coordinate sequence, will produce the salient regions of the input photo. Therefore, 2D spatial attention mechanism is the \emph{key} that connects sequential sketch vector to spatial input photo to predict the saliency map over the input photo.

\keypoint{Sketch Vector Normalisation}
Absolute coordinate based sketch-vector representation is \emph{scale dependent} as the user can draw (see  Fig.~\ref{example_fig}) the same concept in varying scales. Therefore, taking inspiration from sketch/handwriting generation literature \cite{ha2017neural}, we define our sketch-coordinate in terms of \emph{offset} values to make it scale-agnostic. In particular, instead of three elements with absolute coordinate ($x_i,y_i$) and stroke token ($b_i$), now we represent every point as a five-element vector $(\Delta x_i, \Delta y_i, p_i^1, p^2_i,$ $ p^3_i)$, where  $\Delta x_i = (x_{i+1} - x_{i})$ and $\Delta y_i = (y_{i+1} - y_{i})$. Consequently, ($p_i^1, p^2_i, p^3_i$) represents three pen-state situations: pen touching the paper, pen being lifted and end of drawing.



\keypoint{Convolutional Encoder}  Instead of any complicated backbone architectures \cite{szegedy2016rethinking, he2016deep}, we use a straight-forward VGG-16 as the backbone convolutional encoder, which takes a photo (image) $P$ $\in$ $\mathbb{R}^{H \times W \times 3}$ as input and outputs multi-scale convolutional feature  maps as $\mathcal{F}$ $\in$ $\{\mathcal{F}^{l}, \mathcal{F}^{l-1}, \mathcal{F}^{l-2}\}$, where each feature map has a spatial down-scaling factor of $\{32, 16, 8\}$ compared to the input spatial size with $\{512, 512, 256\}$ channels, respectively. This multi-scale design is in line with existing saliency detection literature \cite{piao2019depth}. $\mathcal{F}$ is later accessed by the sequential sketch decoder via an attention mechanism. Unlike multi-modal sketch-synthesis~\cite{ha2017neural, bhunia2021more}, we purposefully ignored the variational formulation~\cite{ha2017neural} as we use sketch coordinate decoding as an auxiliary task \cite{sketch2vec} to generate saliency map from sketch labels.

\keypoint{Sequential Decoder} Our sequential RNN decoder \cite{ha2017neural} uses the encoded feature representation of a photo $\mathcal{F}$ from convolutional encoder to predict a sequence of coordinates and pen-states, which when rasterised, would yield the sketch counterpart of the original photo.  In order to model the variability towards free flow nature of sketch drawing, each off-set position $( \Delta x, \Delta y)$  is modelled using a Gaussian Mixture Model (GMM) with $M = 20$ bivariate normal distributions \cite{ha2017neural} given by $p(\Delta x, \Delta y) = \sum_{j=1}^{M}\Pi_{j}\mathcal{N}(\Delta x, \Delta y | \lambda_{j})$.
$\mathcal{N}(\Delta x, \Delta y | \lambda_{j})$ is {the} probability distribution function for a bivariate normal distribution. Each of $M$ bivariate normal distributions is parameterised by five elements $\lambda = (\mu_x, \mu_y, \sigma_x, \sigma_y, \rho_{xy})$ with mean $(\mu_x, \mu_y)$, standard deviation $(\sigma_x, \sigma_y)$ and correlation $(\rho_{xy})$. While $\mathrm{exp}$ operation is used to ensure non-negative values of mean and standard deviation, $\mathrm{tanh}$ ensures the correlation value is between $-1$ and $1$. The mixture of weights of the GMM  ($\Pi \in \mathbb{R}^{M}$) is modelled by softmax normalised categorical distribution as $\sum_{j}^{M} \Pi_j = 1$. Furthermore, three bit pen state is also modelled by a categorical distribution. Therefore, every time step’s output $y_t$ modelled is of size $\mathbb{R}^{5M+M+3}$, which includes $3$ logits for pen-state.


We perform a global average pooling on $\mathcal{F}^L$ to obtain a feature vector $f_g \in \mathbb{R}^{d}$ which is then fed through a linear embedding layer producing an intermediate latent vector that will be used to initialise the hidden state $h_0$  of decoder RNN as $h_0 = \mathrm{tanh}(W_k f_g + b_k)$. At a time step $t$, the update rule of the decoder network involves a recurrent function of its previous state $s_{t-1}$, and an input concatenating a context vector $g_t$ and coordinate vector $V_{t-1}$ (five elements) obtained in the previous time step as $s_t$ = $(h_t,c_t)$ = $RNN(s_{t-1};[g_t, V_{t-1}])$,
where $[\cdot~; \cdot]$ denotes concatenation and start token $V_{0}$ = $(0, 0, 1, 0, 0)$. The context vector $g_t$ is obtained from a Multiscale 2D attention module, as described later. A fully connected layer over every time step outputs $y_t = W_y h_t + b_y$ where $y_t \in \mathbb{R}^{6M+3}$, which is used to sample five-element coordinate vector using GMM with bivariate normal distribution for offsets and categorical distribution for pen-states as described before.

\keypoint{Multi-Scale 2D Attention Module} This is the \emph{key} module using which we generate the \emph{saliency map} from our photo-to-sketch generation process. Compared to global average pooled  fixed vector for image representation, we design a multiscale-2D attention module through which the sequential decoder looks back at a specific part of the photo at every time step of decoding which it draws. Conditioned on the current hidden state $s_{t-1}$, we first compute $s_{t-1}$-informed feature map as $\mathcal{B}^k_t = W_{M} \circledast  \mathcal{F}^k + \hat{W}_{s}s_{t-1}$ where $k \in \{l, \; l-1, \; l-2\}$, `$\circledast$' is employed with $3 \times 3$ convolutional kernel $W_M$, and $\hat{W}_{s}$ is a learnable weight vector. Next, we combine the individual feature map via simple summation as $\mathcal{B}_t = \mathcal{B}^l_t + \texttt{BD}_{\times2}(\mathcal{B}^{l-1}_t) + \texttt{BD}_{\times4}(\mathcal{B}^{l-2}_t)$, and spatial size is matched using bilinear downscaling ($\texttt{BD}$) operation for $\mathcal{B}^{l-1}_t$  and $\mathcal{B}^{l-2}_t$.  Thereafter, we use the multi-scale aggregated convolutional feature map $\mathcal{B}_t \in \mathbb{R}^{h \times w \times d}$ (with $h = \frac{H}{32}, w = \frac{W}{32}$) where each spatial position in $\mathcal{B}_t$ represents a localised part in original input photo, to get a context vector $g_t$ that represents the most relevant part of $\mathcal{B}_t$ (later $t$ is ignored for brevity) for predicting $y_t$, as:

\vspace{-0.6cm}
\begin{align}
& J =  \mathrm{tanh}(W_{F}\mathcal{B}_{i,j} + W_{B} \circledast  \mathcal{B} + W_{s}s_{t-1}) \nonumber \\
& \alpha_{i,j}^{t}  =  \mathrm{softmax}(W_{a}^T  J_{i,j}) \label{eqn1} \\
& g_{t} =  \sum_{i,j} \alpha_{i,j}^{t} \cdot \mathcal{B}_{i,j}  \; \; i = [1, .., h],  \; j = [1, .., w] \label{eqn2}
\end{align}
\vspace{-0.5cm}

\noindent where,  $W_{F}$, $W_{B}$, $W_{s}$ and $W_{a}$ are learnable weights. While calculating the attention weight $\alpha_{i,j}^{t}$  at every spatial position $(i,j)$, a convolutional operation ``$\circledast$'' is employed with $3 \times 3$ kernel $W_B$ to gather the  neighbourhood information, as well as $1 \times 1$ kernel $W_F$  to consider the local information. We get $\alpha^t \in \mathbb{R}^{h \times w}$ at every time step where every spatial position $\alpha^t(i,j)$ represents the importance on the input image space to draw the next coordinate point. Therefore, given maximum $T$ time steps of sequential sketch coordinate decoding, accumulating over $\alpha^t$ over the whole sequence gives the saliency map as follows:

\vspace{-0.6cm}
\begin{equation}
    \hat{s}_m(i,j) =  \frac{1}{T}\sum_{i=1}^{T} \alpha^t(i,j)\;   \rightarrow \; {s_m}(i,j)  = \frac{\hat{s}_m(i,j)}{max(s_m)}
\end{equation}
\vspace{-0.5cm}

\noindent where, $\hat{s}_m(i,j)$ is un-normalised, but gets normalised later to ensure that every positional scalar saliency value  ${s_m}(i,j)$ lies in the range of $[0,1]$. ${s_m}$ is of size $h \times w$ which can be resized to input photo size  $H \times W$ using bi-linear interpolation to get the final saliency map $S_M$.

\vspace{0.3cm}
\begin{figure}[!htbp]
\includegraphics[width=1.0\linewidth]{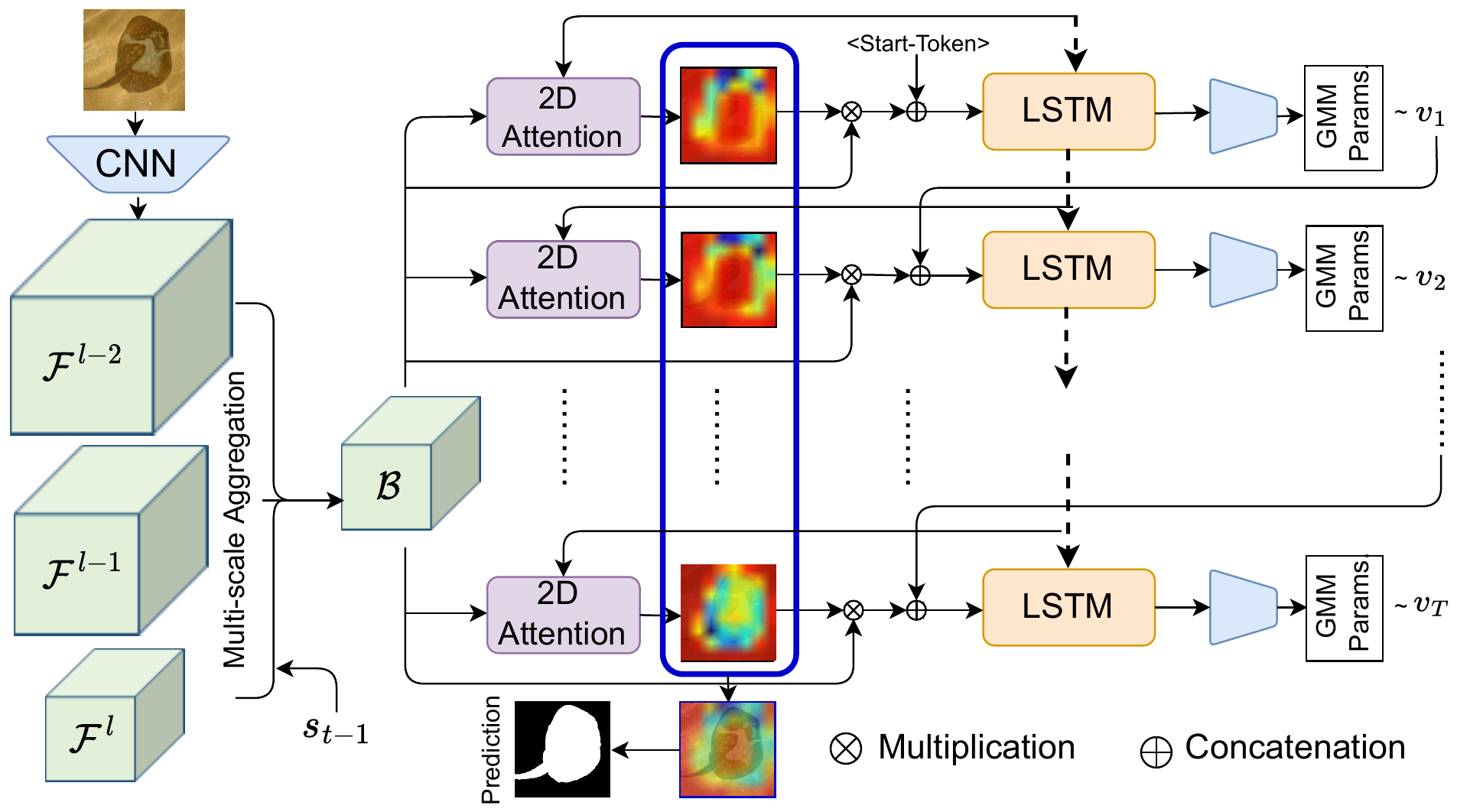}
\vspace{-0.6cm}
\caption{Illustration of photo-to-sketch generation process to learn image saliency from sketch labels. Attention maps accumulated across the time steps give rise to the saliency map.}
\label{fig2}
\vspace{-0.4cm}
\end{figure}

\keypoint{Equivariant Regularisation} As there is a large supervision gap between pixel-level fully supervised labels \cite{zhang2018progressive} and weakly supervised sketch labels, we take inspiration from self-supervised learning \cite{hung2019scops} literature to bring additional inductive bias via auxiliary supervision. In particular, the assumption behind equivariant regularisation is that the output space saliency map $s_m$ should undergo \emph{same} geometric (affine) transformations like that of input image. As the model is aware of pixel-level labelling in a fully supervised setup, the affine transformation consistency is taken care of intrinsically. However, such a protection is absent in weak supervision. There is thus a potential risk of disparity between the generated attention maps and the original photos in case of rotation, scaling or axial flips, leading to inaccurate saliency findings. To alleviate this, we introduce equivariant regularisation, ensuring that attention maps produced, are consistent with the orientation of the original image passed into the encoder network. In general, equivariant regularisation may be defined as:

\vspace{-0.3cm}
\begin{equation}\label{eq:eqv_reg}
    \mathcal{R}_\text{eqv} =  \left \| {\mathcal{X}(\mathcal{A}(P))-\mathcal{A}(\mathcal{X}(P))}  \right \|_{1}
\end{equation}
\vspace{-0.5cm}  

Here,  $\mathcal{A(\cdot)}$ denotes any affine transformation applied on the image, such as axial flipping, rescaling and rotation. $\mathcal{X}(\cdot)$ denotes a encased operation to output the saliency map $S_M$ from input photo $P$ by accumulating attention maps across the sketch vector sequence.

\subsection{Training Objective}

\noindent Our photo-to-sketch generation model (Fig.~\ref{fig2}) is trained in an end-to-end manner with three losses as follows:


\keypoint{Pen State Loss:}
The coordinate-wise pen-state is given by $(\text{\^p}^{1},\text{\^p}^{2},\text{\^p}^{3})$ which is a one-hot vector and hence, optimised as a 3-class softmax classification task by using a categorical cross-entropy loss over a sequence of length $T$:
\vspace{-0.3cm}
\begin{equation}
    \mathcal{L}_\text{coord} = -\frac{1}{T} \sum_{t=1}^{T}\sum_{c=1}^{3}p_{t}^{c}\log{(\text{\^p}_{t}^{c})}
\end{equation}
\vspace{-0.5cm}

\keypoint{Stroke Loss:}
The stroke loss minimises the negative log-likelihood of spatial offsets of each ground truth stroke given the predicted GMM distribution, parameterised by $(\Pi_1, \cdots \Pi_M)$ and $(\lambda_1, \cdots \lambda_M)$, at each time step $t$.
\vspace{-0.3cm}
\begin{equation}
\label{eq6}
    \mathcal{L}_\text{stroke} = -\frac{1}{T}\sum_{t=1}^{T} \log \big(\sum_{j=1}^{M} \Pi_{t,j} \mathcal{N}(\Delta x_t, \Delta y_t | \lambda_{t,j})\big)
\end{equation}
\vspace{-0.5cm}

\keypoint{Equivariant Loss:}
The equivariant regularisation loss aims to preserve spatial consistency between the input photo $P$ and the generated saliency map $S_M$, viz:

\vspace{-0.3cm}
\begin{equation}
    \mathcal{L}_\text{eqv} = \left \|  {\mathcal{X}(\mathcal{A}(P)) - \mathcal{A}(S_M)} \right \|_{1}   
\end{equation}
\vspace{-0.5cm}

Here, $\mathcal{A}$ is any affine transformation (rotation, flipping, scaling) and $\mathcal{X}(\cdot)$ denotes an encased operation to output saliency map $S_M$. Overall, $\mathcal{L}_\text{sketch} = \mathcal{L}_\text{coord} + \mathcal{L}_\text{stroke} + \mathcal{L}_\text{eqv}$.


\vspace{-0.2cm}
\section{Experiments}\label{sec:experiments}
\vspace{-0.2cm}
\keypoint{Datasets:} To train our saliency detection framework from sketch labels, we use Sketchy dataset \cite{sangkloy2016sketchy}  which contains photo-sketch pairs of $125$ classes. Each  photo  has  at  least $5$ sketches with fine-grained associations. Altogether, there are $75,471$ sketches representing $12,500$ photos. Ramer-Douglas-Peucker (RDP) algorithm \cite{visvalingam1990douglas} is applied to simplify the sketches to ease the training of LSTM based sequential decoder. Furthermore, we evaluate our model in six benchmark datasets: ECSSD \cite{yan2013hierarchical}, DUT-OMRON\cite{yang2013saliency}, PASCAL-S \cite{li2014secrets}, PASCAL VOC2010\cite{everingham2010pascal}, MSRA5K \cite{liu2010learning}, SOD \cite{martin2001database}. While ECSSD dataset \cite{yan2013hierarchical} consists of 1000 natural images with numerous objects of different orientations from the web, DUT-OMRON \cite{yang2013saliency} contains a total of $5168$ challenging images with one or more salient objects placed in complex backgrounds. PASCAL-S \cite{li2014secrets} incorporates a total of $850$ natural images which is the subset (validation set) of PASCAL VOC2010 \cite{everingham2010pascal} segmentation challenge. MSRA5K \cite{liu2010learning} holds a collection of $5,000$ natural images. SOD \cite{martin2001database} contains 300 images, mainly created for image segmentation; \cite{jiang2013salient} takes the initiative to have the pixel wise annotation of the salient objects.

\vspace{-0.2cm}
\begin{figure}[!hbt]
\begin{center}
  \includegraphics[width=1\linewidth]{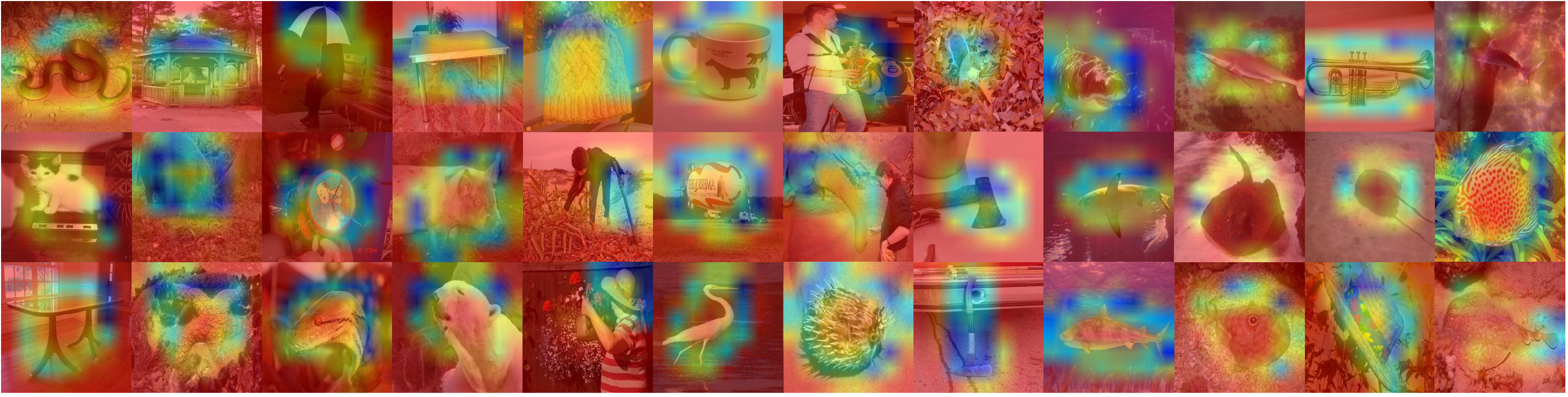}
\end{center}
\vspace{-0.7cm}
  \caption{Visualisation of raw attention-map obtained via sketch as a weak label for saliency detection. While most existing weakly supervised methods involve heuristic-based post-processing techniques \cite{zeng2019multi, liu2021weakly}, all results shown here are directly predicted by the network without any post-processing.}
\label{fig45}
\end{figure}
\vspace{-0.4cm}

\keypoint{Implementation Details:} Implementing our framework in PyTorch \cite{paszke2017automatic}, we conducted experiments on a 11-GB Nvidia RTX 2080-Ti GPU. While an ImageNet pre-trained VGG-16 is used as the backbone convolutional feature extractor for input images, the sequential decoder uses a single layer LSTM with hidden state size of $512$, and embedding dimension for 2D-attention module as $256$. Maximum sequence length is set to $250$. We use Adam optimiser with learning rate $10^{-4}$ and batch size $16$ for training our model for $50$ epochs. Every image is resized to $256 \times 256$ before training. We do not use any heuristic-based post-processing (e.g. CRF \cite{zhang2020weakly}) for evaluation. Importantly, while applying affine-transformations on input photo space for equivariant regularisation, we need to ensure that the \textit{same} transformation is applied on the sketch vector labels as well.

\keypoint{Evaluation Metrics:} We use two evaluation metrics \cite{zeng2019multi} to evaluate our method against other state-of-the-arts: (a) Mean Absolute Error (MAE) -- smaller is better, and (b) maximum F-measure (max $F_\beta$) -- larger is better,  where  $\beta^{2}$ is set to 0.3 \cite{zeng2019multi}. Later on, we compare with recently introduced metrics \cite{wu2019stacked} like weighted F-measure $(F_\beta^w)$ and structural similarity measure ($S_\alpha$; $\alpha = 0.5$).

\vspace{-0.1cm}
\subsection{Performance Analysis}
\vspace{-0.2cm}
\keypoint{Comparison with Alternative Label Source} In weakly supervised saliency detection, various labelling sources have been explored to help reduce cost and time of pixel-wise labelling. While \cite{zeng2019multi} uses textual descriptions for saliency learning, \cite{wang2017learning} employs \emph{class-label} as weak supervision, the hypothesis being that the network should have \emph{high activation} at visually salient foreground regions. However, an important point to note is that most of these methods \cite{liu2021weakly, zhang2020weakly, tian2020weakly} apply heuristic-based \emph{post-processing} on the initial saliency maps to get pixel-level pseudo ground-truth maps to be used by another saliency detection network during training. This dilutes the contribution of the original label source and puts more weight on the \emph{post-processing} heuristic for final performance instead. For instance, while \cite{liu2021weakly} leverages several iterative refinements using multi-stage training to obtain the pseudo ground-truth, scribbles \cite{zhang2020weakly} need an auxiliary edge-detection network for training. Constrastingly, in the \emph{first phase} of our experiments, we focus on validating the \emph{un-augmented} (see Fig. \ref{fig45}) potential of individual label sources that
{comes out directly after end-to-end training.} 
Thus, among alternative label sources, we compare with class-label and text-caption, as these means allow direct usage of saliency maps from end-to-end training without any post-processing. The rest \cite{liu2021weakly, zhang2020weakly, tian2020weakly} are designed such that it is necessary to have some heuristic post-processing step. We also compare with such weakly supervised alternatives separately in \cref{sec:MS-WSSD}.

Table \ref{tab:my_label2} shows comparative results on ECSSD and PASCAL-S datasets. Overall, sketch has a significant edge over class-label and text-captions on both datasets. While class-label is \emph{too crude an information}, text-caption contains many irrelevant details in the description such that not every word in the caption directly corresponds to salient object. In contrast, the superiority of our framework using sketch labels is attributed to two major factors -- (a) Sketch inherently depicts the shape morphology of corresponding salient objects from the paired photo. (b) Our equivariant regularisation further imposes additional auxiliary supervision to learn the inductive bias via weak supervision. Although there are other weakly-supervised methods \cite{yu2021structure, piao2021critical, zhang2020weakly}, we intentionally exclude them from Tables \ref{tab:my_label2} and \ref{tab:my_label4}, as they need heuristic-based pre-/post-processing steps, whereas  we validate the un-augmented potential of sketch for weak labels, as compared in \cref{sec:MS-WSSD}.

\vspace{-0.3cm}
\begin{table}[!htbp]
  \centering
  \small
  \setlength{\tabcolsep}{2pt}
  \renewcommand{\arraystretch}{0.8}
  \caption{Evaluation (no post-processing)}
  \vspace{-0.3cm}
  \begin{tabular}{ccccc}
    \toprule
    \multirow{2}{*}{Method}  & \multicolumn{2}{c}{ECSSD} & \multicolumn{2}{c}{PASCAL-S} \\
    \cmidrule(lr){2-3}\cmidrule(lr){4-5}
                                      & max $F_{\beta}\uparrow$ & MAE$\downarrow$ & max $F_{\beta}\uparrow$ & MAE$\downarrow$ \\
    \cmidrule(lr){1-3}\cmidrule(lr){4-5}
    Class-Label \cite{zeng2019multi}  & 0.720                   & 0.213           & 0.623                   & 0.275           \\
    Text-Caption \cite{zeng2019multi} & 0.730                   & 0.194           & 0.641                   & 0.264           \\
    \rowcolor{Gray}
    Sketch (Proposed)                 & 0.781                   & 0.152           & 0.705                   & 0.228           \\
    \bottomrule
  \end{tabular}
  \label{tab:my_label2}
\end{table}
\vspace{-0.3cm}

\keypoint{Ablative Study:} We have conducted a thorough ablative study on ECSSD dataset to delineate the contributions of different modules present in our framework. (i) Removing \emph{sketch vector normalisation}, and using absolute sketch coordinate drops the max $F_{\beta}$  value by a huge margin of $0.341$, thus signifying its utility behind achieving scale-invariance for sketch vectors. (ii) Instead of using $3 \times 3$ kernel to aggregate the \emph{neighbourhood information} inside the 2D attention module, using simpler 1D attention that considers convolutional feature-map as a sequence of vectors deteriorates the max $F_{\beta}$  to $0.763$ from our final value of $0.781$. (iii) Furthermore, removing \emph{equivariance loss} (Eqn. \ref{eq:eqv_reg}) degrades the performance by a significant margin of $0.021$ max $F_{\beta}$ value. Therefore, it affirms that self-supervised equivariance constraints our weakly supervised model in learning invariance against a variety of perspective deformations common in wild scenarios. (iv) While learning spatial attention-based saliency from class-labels or text-caption could be attained in a single feed-forward pass, ours require sequential auto-regressive decoding, thus increasing the time complexity of our framework. Overall, it needs a CPU time of {32.3 ms} to predict the saliency map of a single input image. However, the performance gain within reasonable remit of time complexity and ease of collecting amateur sketches without any expert knowledge makes sketch a potential label-alternative for saliency detection. (v) Varying \emph{number of Gaussians} M in GMM from $5$ to $25$ at an interval of $5$, we get the following max $F_{\beta}$ values: $0.741$, $0.756$, $0.771$ $0.781$, and $0.779$ respectively, thus optimally choosing $M=20$. (vi) Removing multi-scale feature-map aggregation design reduce the max $F_{\beta}$ by $0.013$. (v)
Overall, removing all our careful design choices and using simple spatial attention at the end of convolutional encoder as per the usual saliency literature  \cite{zeng2019multi} does not fully incorporate the sketching process, and an  experiment sees max $F_{\beta}$ falling to $0.718$ which is even lower than that of the class/text-captions, thus validating the utility of our careful design choices.

{\setlength{\tabcolsep}{2pt}
\renewcommand{\arraystretch}{0.7}
\begin{table}[!htbp]
\centering
\scriptsize
\vspace{-0.2cm}
\caption{{Linear model evaluation using fixed pre-trained features, and Semi-supervised fine-tuning from 1\% and 10\% labelled training data from DUTS\cite{wang2017learning} and evaluated on ECSSD \cite{yan2013hierarchical}.}}
\vspace{-0.3cm}
    \begin{tabular}{cccccccc}
        \toprule
        \multirow{3}{*}{Method} & \multirow{3}{*}{Conv}
        & \multicolumn{2}{c}{Linear Evaluation}
        & \multicolumn{4}{c}{Semi-supervised Evaluation} \\
        \cmidrule(lr){3-4}\cmidrule(lr){5-8}
        & & \multirow{2}{*}{max $F_{\beta}\uparrow$} & \multirow{2}{*}{MAE $\downarrow$} &\multicolumn{2}{c}{1\% Train Data} & \multicolumn{2}{c}{10\% Train Data}  \\
        & & & & max $F_{\beta}\uparrow$ & MAE$\downarrow$ & max $F_{\beta}\uparrow$ & MAE$\downarrow$\\
        \cmidrule(lr){1-4}\cmidrule(lr){5-6}\cmidrule(lr){7-8}
        Fully \cite{wang2015deep} & {$1\times 1$} & 0.869 & 0.118  & 0.708 & 0.264 & 0.753 & 0.233  \\
        Supervised & {$3 \times 3$} & 0.872 & 0.111 & 0.713 & 0.251 & 0.761 & 0.224\\
        \cmidrule(lr){1-4}\cmidrule(lr){5-6}\cmidrule(lr){7-8}
        Class \cite{zeng2019multi} &{$1\times 1$} & 0.781 & 0.172 & 0.731 & 0.210 & 0.757 & 0.191\\
        Label &{$3 \times 3$} & 0.784 & 0.167 & 0.735 & 0.207 & 0.763 & 0.186 \\
        \cmidrule(lr){1-4}\cmidrule(lr){5-6}\cmidrule(lr){7-8}
        Text \cite{zeng2019multi} &{$1\times 1$} & 0.789 & 0.161 & 0.738 & 0.191 & 0.765 & 0.180 \\
        Caption &{$3 \times 3$} & 0.793 & 0.153 & 0.742 & 0.187 & 0.771 & 0.172\\
        \cmidrule(lr){1-4}\cmidrule(lr){5-6}\cmidrule(lr){7-8}
        \rowcolor{Gray}
         & {$1\times 1$} & 0.832 & 0.129 & 0.788 & 0.148 & 0.817 & 0.145\\
         \rowcolor{Gray}
        \multirow{-2}{*}{Sketch} & {$3 \times 3$} & 0.841 & 0.121 & 0.791 & 0.142 & 0.822 & 0.138\\
       \bottomrule
    \end{tabular}
    \vspace{-0.2cm}
    \label{tab:my_label4}
\end{table}
}

\begin{figure*}[!hbt]
\begin{center}
  \includegraphics[width=\linewidth]{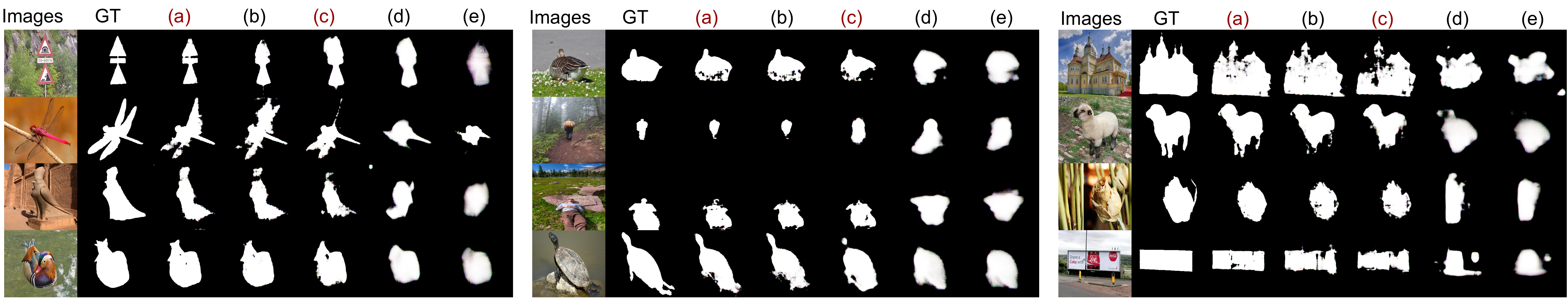}
\end{center}
\vspace{-0.6cm}
  \caption{Qualitative results on weakly supervised saliency detection using (a) class-label + sketch  (b) class-label + text-caption \cite{zeng2019multi} (c) only sketch  (d) only text-caption  (e) only class-label. Use of sketch over text-caption significantly improves the quality of saliency map .}
  \vspace{-0.6cm}
  \label{fig:qual}
\end{figure*}

\vspace{-0.2cm}
\subsection{Photo-to-Sketch as Pre-training for Saliency}
\vspace{-0.05cm}
Following the self-supervised literature \cite{jing2020self, sketch2vec}, we further aim to judge the goodness of the learned convolutional feature map from our photo-to-sketch generation task for saliency detection in general. In particular, having trained the photo-to-sketch generation model from scratch, we remove the components of sequential decoder and freeze the weights of convolutional encoder, which thus acts as a standalone feature extractor.
Following the linear evaluation protocol of self-supervised learning \cite{jing2020self}, we resize the last output feature map to the size of input image using bilinear interpolation, and apply \emph{one} learnable layer for pixel-wise binary classification. We compare with  $1\times1$ and $3\times3$ convolutional kernel for that learnable layer which is trained using DUTS dataset \cite{wang2017learning} containing $10,533$ training images with pixel-level labels using binary-cross entropy loss. The results in Table \ref{tab:my_label4} (left) shows the primacy of sketch based weakly supervised pre-training over others, illustrating the superior generalizability of the sketch learned features over others for saliency detection. {We next evaluate the semi-supervised setup, where we fine-tune the whole network \cite{jing2020self} using smaller subsets of the training data (i.e., $1\%$ and $10\%$). Our photo-to-sketch generation using weak sketch labels as pre-training strategy augments initialisation such that even in this low data regime, our approach outperforms its supervised training counterpart under the same architecture as shown in Table~\ref{tab:my_label4} (right).}


\setlength{\tabcolsep}{2.5pt}
\renewcommand{\arraystretch}{0.6}
\begin{table*}[!hbt]
\centering
\caption{Comparative study with SOTA saliency detection methods on ECSSD \cite{yan2013hierarchical}, PASCAL-S \cite{li2014secrets}, MSRA5K \cite{liu2010learning}, SOC \cite{fan2018salient} dataset.}
\label{tab:my_label6}
\vspace{-0.3cm}
\scriptsize
\begin{tabular}{clcccccccccccccccc}
\toprule
\multirow{2}{*}{Type} & \multirow{2}{*}{Method}   & \multicolumn{4}{c}{ECSSD}                                                            & \multicolumn{4}{c}{PASCAL-S}                                                         & \multicolumn{4}{c}{MSRA5K}  & \multicolumn{4}{c}{SOC}                                                          \\
\cmidrule(lr){3-6}\cmidrule(lr){7-10}\cmidrule(lr){11-14}\cmidrule(lr){15-18}
                      &
                      & max $F_{\beta}$$\uparrow$ & MAE $\downarrow$ & $F_\beta^w$$\uparrow$  & $S_\alpha$$\uparrow$
                      & max $F_{\beta}$$\uparrow$ & MAE $\downarrow$ & $F_\beta^w$$\uparrow$ & $S_\alpha^w$$\uparrow$
                      & max $F_{\beta}$$\uparrow$ & MAE $\downarrow$ & $F_\beta^w$$\uparrow$  & $S_\alpha^w$$\uparrow$
                      & max $F_{\beta}$$\uparrow$ & MAE $\downarrow$ & $F_\beta^w$$\uparrow$  & $S_\alpha^w$$\uparrow$                       \\
\cmidrule(lr){1-6}\cmidrule(lr){7-10}\cmidrule(lr){11-14}\cmidrule(lr){15-18}

\multirow{2}{*}{FS}   & {LEGS \cite{wang2015deep}}                      & 0.827         & 0.118            & -- & \textcolor[rgb]{0.141,0.133,0.118}{--} & 0.761         & 0.155            & -- & \textcolor[rgb]{0.141,0.133,0.118}{--} & –             & –                & -- & \textcolor[rgb]{0.141,0.133,0.118}{--}  & -- & -- & -- & --\\

                      & {DS \cite{li2016deepsaliency}}                        & {0.882}         & 0.122            & -- & \textcolor[rgb]{0.141,0.133,0.118}{--} & 0.763         & 0.176            & -- & \textcolor[rgb]{0.141,0.133,0.118}{--} & –             & –                & -- & \textcolor[rgb]{0.141,0.133,0.118}{--}  & -- & -- & -- & --\\
\cmidrule(lr){1-6}\cmidrule(lr){7-10}\cmidrule(lr){11-14}\cmidrule(lr){15-18}
\multirow{3}{*}{US}   & {BSCA \cite{qin2015saliency}}                      & 0.758         & 0.182            & -- & \textcolor[rgb]{0.141,0.133,0.118}{--} & 0.663         & 0.223            & -- & \textcolor[rgb]{0.141,0.133,0.118}{--} & 0.829         & 0.132            & -- & \textcolor[rgb]{0.141,0.133,0.118}{--}  & -- & -- & -- & --\\
                      & {MB \cite{zhang2015minimum}}                        & 0.736         & 0.193            & -- & \textcolor[rgb]{0.141,0.133,0.118}{--} & 0.673         & 0.228            & -- & \textcolor[rgb]{0.141,0.133,0.118}{--} & 0.822         & 0.133            & -- & \textcolor[rgb]{0.141,0.133,0.118}{--}  & -- & -- & -- & --\\
                      & {MST \cite{tu2016real}}                       & 0.724         & 0.155            & -- & \textcolor[rgb]{0.141,0.133,0.118}{--} & 0.657         & 0.194            & -- & \textcolor[rgb]{0.141,0.133,0.118}{--} & 0.809         & 0.980             & -- & \textcolor[rgb]{0.141,0.133,0.118}{--}  & -- & -- & -- & --\\
\cmidrule(lr){1-6}\cmidrule(lr){7-10}\cmidrule(lr){11-14}\cmidrule(lr){15-18}
\multirow{8}{*}{WS}   & {Cls-Label \cite{wang2017learning}}                 & 0.856         & 0.104            & -- & \textcolor[rgb]{0.141,0.133,0.118}{--} & 0.778         & 0.141            & -- & \textcolor[rgb]{0.141,0.133,0.118}{--} & 0.877         & 0.076            & -- & \textcolor[rgb]{0.141,0.133,0.118}{--}  & -- & -- & -- & --\\
                      & {Cls-Label \cite{zeng2019multi}}                 & 0.773         & 0.162            & 0.721 & \textcolor[rgb]{0.141,0.133,0.118}{0.753} & 0.736         & 0.176            & 0.672& \textcolor[rgb]{0.141,0.133,0.118}{0.734} & 0.836         & 0.094            & 0.812 & \textcolor[rgb]{0.141,0.133,0.118}{0.842}  & \textcolor[rgb]{0.145,0.137,0.118}{0.581} & \textcolor[rgb]{0.145,0.137,0.118}{0.167} & \textcolor[rgb]{0.145,0.137,0.118}{0.521} & \textcolor[rgb]{0.145,0.137,0.118}{0.657}\\
                      & {Text-Caption \cite{zeng2019multi}}              & 0.796         & 0.152            & 0.741 & \textcolor[rgb]{0.141,0.133,0.118}{0.773} & 0.748         & 0.167            & 0.681 & \textcolor[rgb]{0.141,0.133,0.118}{0.743} & 0.848         & 0.087            & 0.823 & \textcolor[rgb]{0.141,0.133,0.118}{0.851}  & \textcolor[rgb]{0.145,0.137,0.118}{0.593} & \textcolor[rgb]{0.145,0.137,0.118}{0.158} & \textcolor[rgb]{0.145,0.137,0.118}{0.543} & \textcolor[rgb]{0.145,0.137,0.118}{0.661}\\

                    & {Bounding-box \cite{liu2021weakly}}              & 0.860        & 0.072            & 0.820 & \textcolor[rgb]{0.141,0.133,0.118}{0.858} & --         & --            & -- & \textcolor[rgb]{0.141,0.133,0.118}{--} & --         & --            & -- & \textcolor[rgb]{0.141,0.133,0.118}{--}  & -- & -- & -- & --\\

                    & {Scribble \cite{zhang2020weakly}}              & 0.865        & \textbf{0.061}            & 0.823 & \textcolor[rgb]{0.141,0.133,0.118}{0.867} & 0.788         & 0.139            & \textbf{0.753} & \textbf{0.779} & --         & --            & -- & \textcolor[rgb]{0.141,0.133,0.118}{--}  & -- & -- & -- & --\\

                    & {Class-Label \cite{piao2021critical}}              & 0.853        & 0.083            & -- & \textcolor[rgb]{0.141,0.133,0.118}{--} & 0.713         & 0.133            & -- & \textcolor[rgb]{0.141,0.133,0.118}{--} & --         & --            & -- & \textcolor[rgb]{0.141,0.133,0.118}{--}  & -- & -- & -- & --\\

                  & {Scribble \cite{yu2021structure}}              & 0.865        & \textbf{0.061}            & 0.845 & \textcolor[rgb]{0.141,0.133,0.118}{\textbf{0.881}} & 0.758         & 0.684            & 0.732 & \textcolor[rgb]{0.141,0.133,0.118}{0.761} & --         & --            & -- & \textcolor[rgb]{0.141,0.133,0.118}{--}  & -- & -- & -- & --\\

                      & {Cls+Text \cite{zeng2019multi}}                  & 0.878         & 0.096            & \textbf{0.857} & \textcolor[rgb]{0.141,0.133,0.118}{0.876} & 0.790         & 0.134            & 0.773 & \textcolor[rgb]{0.141,0.133,0.118}{0.763} & 0.890         & 0.071            & 0.872 & \textcolor[rgb]{0.141,0.133,0.118}{0.869}  & \textcolor[rgb]{0.145,0.137,0.118}{0.639} & \textcolor[rgb]{0.145,0.137,0.118}{0.131} & \textcolor[rgb]{0.145,0.137,0.118}{0.571} & \textcolor[rgb]{0.145,0.137,0.118}{0.741}\\
\cmidrule(lr){1-6}\cmidrule(lr){7-10}\cmidrule(lr){11-14}\cmidrule(lr){15-18}
\rowcolor{Gray}
 & \textbf{Sketch}           & 0.843         & 0.123            & 0.811 & \textcolor[rgb]{0.141,0.133,0.118}{0.832} & 0.761         & 0.149            & 0.723 & \textcolor[rgb]{0.141,0.133,0.118}{0.754} & 0.867         & 0.089            & 0.832 & \textcolor[rgb]{0.141,0.133,0.118}{0.854}  & \textcolor[rgb]{0.145,0.137,0.118}{0.612} & \textcolor[rgb]{0.145,0.137,0.118}{0.149} & \textcolor[rgb]{0.145,0.137,0.118}{0.552} & \textcolor[rgb]{0.145,0.137,0.118}{0.721}\\
\rowcolor{Gray}
   \multirow{-2}{*}{Ours}                   & \textbf{Cls-Label+Sketch} & \textbf{0.881}         & 0.072            & 0.843 & \textcolor[rgb]{0.141,0.133,0.118}{0.874} & \textbf{0.808}         & \textbf{0.126}            & 0.752 & \textcolor[rgb]{0.141,0.133,0.118}{0.778} & \textbf{0.909}         & \textbf{0.056}            & \textbf{0.883} & \textbf{0.901}  & \textbf{0.651} & \textbf{0.123} & \textbf{0.587} &  \textbf{0.755}\\
\bottomrule
\end{tabular}
\end{table*}


\setlength{\tabcolsep}{2.2pt}
\renewcommand{\arraystretch}{0.6}
\begin{table*}[!hbt]
\label{tab:my_label7}
\vspace{-0.01cm}
\scriptsize
\begin{minipage}{0.48\linewidth}
\caption{Comparision on SOD \cite{martin2001database} and DUT-OMRON \cite{yang2013saliency}.}
\vspace{-0.25cm}
\begin{tabular}{clcccccccc}
\toprule
\multirow{2}{*}{Type} & \multirow{2}{*}{Method} & \multicolumn{4}{c}{SOD} & \multicolumn{4}{c}{DUT-OMRON} \\
\cmidrule(lr){3-6}\cmidrule(lr){7-10}
     &   & max $F_{\beta} \uparrow$ & MAE $\downarrow$ & $F_\beta^w \uparrow$  & $S_\alpha \uparrow$
         & max $F_{\beta} \uparrow$ & MAE $\downarrow$ & $F_\beta^w \uparrow$ & $S_\alpha^w \uparrow$ \\
\cmidrule(lr){1-6}\cmidrule(lr){7-10}
\multirow{2}{*}{FS}   & {LEGS \cite{wang2015deep}}                      & 0.733         & 0.196            & -- & \textcolor[rgb]{0.145,0.137,0.118}{--} & 0.671         & 0.140            & -- & \textcolor[rgb]{0.145,0.137,0.118}{--} \\
                      & {DS \cite{li2016deepsaliency}}                        & 0.784         & 0.190            & -- & \textcolor[rgb]{0.145,0.137,0.118}{--} & 0.739         & 0.127            & -- & \textcolor[rgb]{0.145,0.137,0.118}{--} \\
\cmidrule(lr){1-6}\cmidrule(lr){7-10}
\multirow{3}{*}{US}   & {BSCA \cite{qin2015saliency}}                      & 0.656         & 0.252            & -- & \textcolor[rgb]{0.145,0.137,0.118}{--} & 0.613         & 0.196            & -- & \textcolor[rgb]{0.145,0.137,0.118}{--} \\
                      & {MB \cite{zhang2015minimum}}                        & 0.658         & 0.255            & -- & \textcolor[rgb]{0.145,0.137,0.118}{--} & 0.621         & 0.193            & -- & \textcolor[rgb]{0.145,0.137,0.118}{--} \\
                      & {MST \cite{tu2016real}}                       & 0.647         & 0.223            & -- & \textcolor[rgb]{0.145,0.137,0.118}{--} & 0.588         & 0.161            & -- & \textcolor[rgb]{0.145,0.137,0.118}{--} \\
\cmidrule(lr){1-6}\cmidrule(lr){7-10}
\multirow{8}{*}{WS}   & {Cls-Label \cite{wang2017learning}}                & 0.780         & 0.170            & -- & \textcolor[rgb]{0.145,0.137,0.118}{--} & 0.687         & 0.118            & -- & \textcolor[rgb]{0.145,0.137,0.118}{--} \\
    & {Cls-Label \cite{zeng2019multi}}                 & 0.738         & 0.214            & 0.693 & \textcolor[rgb]{0.145,0.137,0.118}{0.721} & 0.627         & 0.176    & 0.581 & \textcolor[rgb]{0.145,0.137,0.118}{0.619} \\
    & {Text-Caption \cite{zeng2019multi}}              & 0.748         & 0.203            & 0.712 & \textcolor[rgb]{0.145,0.137,0.118}{0.736} & 0.641         & 0.169     & 0.623 & \textcolor[rgb]{0.145,0.137,0.118}{0.656} \\
    & {Bounding-box \cite{liu2021weakly}} & -- & -- & -- & \textcolor[rgb]{0.141,0.133,0.118}{--} & 0.680         & 0.081            & 0.650 & \textbf{0.770} \\
    & {Scribble \cite{zhang2020weakly}}              &--        & --            & -- & \textcolor[rgb]{0.141,0.133,0.118}{--} & 0.701         & \textbf{0.068}            & 0.712 & \textcolor[rgb]{0.141,0.133,0.118}{0.721}  \\
    & {Class-Label \cite{piao2021critical}}              & --        & --            & -- & \textcolor[rgb]{0.141,0.133,0.118}{--} & 0.667         & 0.078            & -- & \textcolor[rgb]{0.141,0.133,0.118}{--} \\
    & {Scribble \cite{yu2021structure}}              &--        & --            & -- & \textcolor[rgb]{0.141,0.133,0.118}{--} & \textbf{0.758}         & 0.069            & \textbf{0.745} & \textcolor[rgb]{0.141,0.133,0.118}{0.754} \\
    & {Cls+Text \cite{zeng2019multi}}                  & 0.799         & 0.167            & 0.756 & \textcolor[rgb]{0.145,0.137,0.118}{0.767} & 0.718         & 0.114            & 0.675 & \textcolor[rgb]{0.145,0.137,0.118}{0.689} \\
\cmidrule(lr){1-6}\cmidrule(lr){7-10}
\rowcolor{Gray}
 & \textbf{Sketch}           & 0.763         & 0.183            & 0.744 & \textcolor[rgb]{0.145,0.137,0.118}{0.768} & 0.673         & 0.132            & 0.653 & \textcolor[rgb]{0.145,0.137,0.118}{0.683} \\
\rowcolor{Gray}
\multirow{-2}{*}{Ours} & \textbf{Cls-Label+Sketch} & \textbf{0.813}         & \textbf{0.151}            & \textbf{0.798} & \textbf{0.788} & 0.726         & 0.101            & 0.701 & \textcolor[rgb]{0.145,0.137,0.118}{0.723} \\
\bottomrule
\end{tabular}
\end{minipage}\hfill
\begin{minipage}{0.42\linewidth}
    \centering
    \includegraphics[width=1\columnwidth, height=3.5cm]{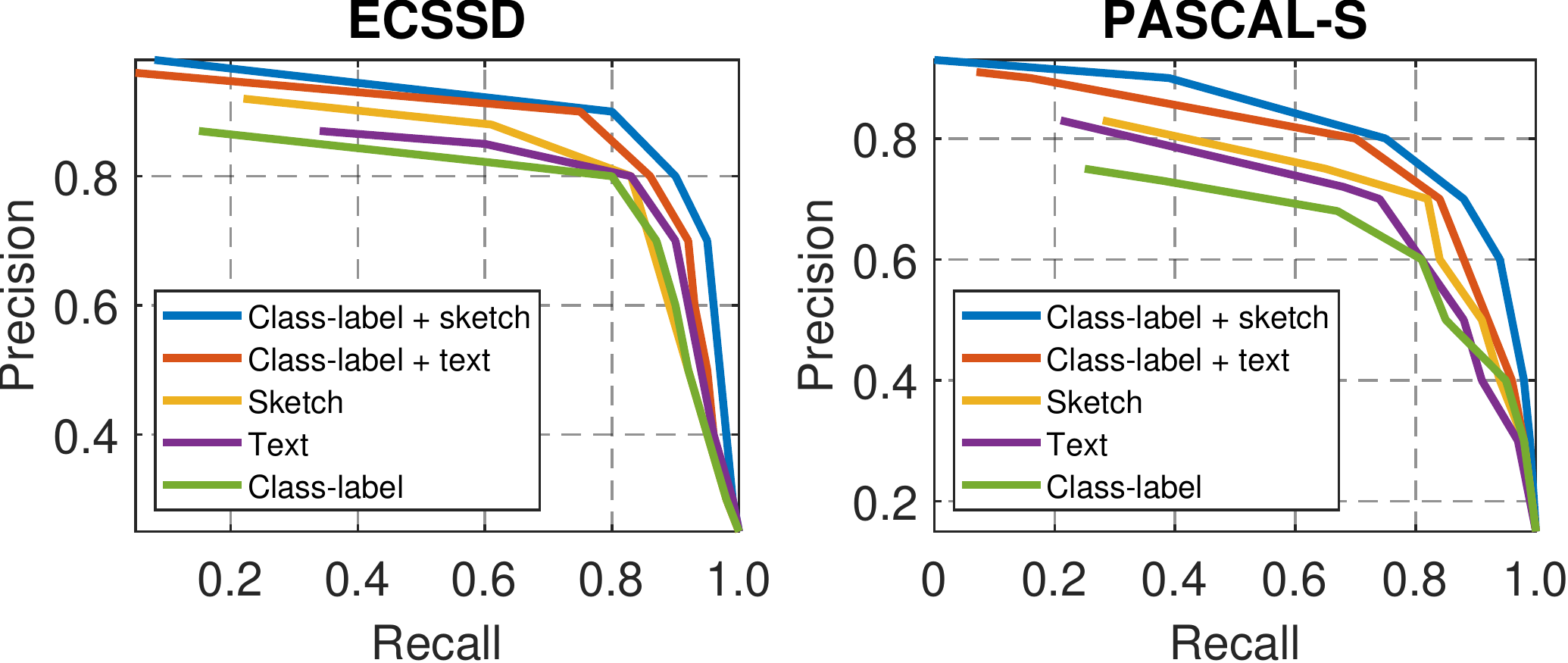}
    \vspace{-0.5cm}
    \captionof{figure}{Precision-Recall curves of models trained from different label source under weakly supervised setup.}
    \label{fig:iitd}
  \end{minipage}
\vspace{-0.3cm}
\end{table*}

\vspace{-0.2cm}
\section{Extend Sketch Labels to MS-WSSD}  \label{ms-wssd}
\label{sec:MS-WSSD}
\vspace{-0.2cm}
\keypoint{Overview:}  This work aims at validating the potential of sketches \cite{hertzmann2020line} as labels to learn visual saliency \cite{li2015visual}. Therefore, as the primary intent of this work, we design a simple and easily reproducible sequential photo-to-sketch generation framework with an additional equivariance objective \cite{wang2020self}. Recently, lacking a single weak supervision source, attempts have been made towards designing a Multi-Source Weakly Supervised Saliency Detection (MS-WSSD) framework \cite{zeng2019multi}. Therefore, we further aim to examine the efficacy of sketches when being incorporated in such an MS-WSSD pipeline without many bells and whistles, and compare with state-of-the-art WSSD frameworks~\cite{wei2017object,oh2017exploiting}.

We ablate the framework by Zheng \etal \cite{zeng2019multi} that leveraged weak supervision via one-hot encoded \emph{class-labels} and \emph{text-descriptions} (captions), replacing sketch as the source of label \emph{instead of} text-caption. We follow a two-stage training: First we train a classification and a photo-to-sketch generation network, which helps generate pseudo-pixel-level ground-truth saliency maps for unlabelled images. Next these pseudo-pixel-level ground-truths help train a separate saliency detection network with supervised loss.

\vspace{0.1cm}
\noindent \textbf{Model:} For the Classification network (C-Net), given an extracted feature-map from VGG-16 network $\mathcal{F} \in \mathbb{R}^{h \times w \times c}$, we apply $1\times1$ convolution followed by sigmoid function to predict the coarse saliency map $s_m^c \in \mathbb{R}^{h \times w }$. Thereafter, $\mathcal{F}$ is multiplied by channel wise extended $s_m^c$ to re-weight the feature-map $\mathcal{F}' = \mathcal{F}\cdot s_m^c$. We apply global average pooling on $\mathcal{F}'$, followed by a linear classification layer to predict the probability distribution over classes. For the photo-to-sketch generation network (S-Net), we use our off-the-shelf network (\cref{sec:method}). The coarse saliency map obtained by S-Net is denoted as  $s_m^s \in \mathbb{R}^{h \times w}$.

\noindent \textbf{Loss Functions:} \emph{Firstly}, we train the C-Net using available class-level training data with cross-entropy loss across class-labels and binary cross-entropy loss on $s_m^c$ with ground-truth matrix of all zeros. This acts as a regulariser to prevent the trivial saliency maps having high responses at all locations, thereby encouraging to focus on the foreground object region important for classification. We term the classification network loss as $\mathcal{L}_\text{class}$. \emph{Secondly}, we train the S-Net model using available data with  $\mathcal{L}_\text{sketch}$. \emph{Thirdly}, we co-train C-Net and S-Net with \emph{two additional} losses -- attention transfer loss and attention coherence loss. Attention transfer loss aims to transfer the knowledge learned from C-Net towards S-Net, and vice-versa. The predicted coarse saliency map by C-Net (thresholded by $0.5$) acts as a ground-truth for S-Net, and vice-versa. Contrarily, the attention coherence loss utilises low-level colour similarity and super-pixels to make the predicted coarse saliency maps from C-Net and S-Net similar. Please refer to \cite{zeng2019multi} for more details on the loss functions.

\noindent \textbf{Post-Processing and Training Saliency Network:} For the second stage of the framework, we use the predicted saliency maps from C-Net and S-Net as ground-truth for unlabelled photos to train a separate saliency detection network. Both the coarse saliency maps $s_m^c$ and $s_m^s$ are averaged and resized to that of original image using bilinear interpolation. The transformed map is processed with CRF~\cite{krahenbuhl2011efficient} during training and binarized to form the pseudo labels, while bootstrapping loss \cite{reed2014training} is used to train the saliency detection network (Sal-N). We use prediction from Sal-N as the final result without any post processing over there, and our results (Tables~\ref{tab:my_label6} and \red{4}) follow the same.

\vspace{-0.1cm}
\subsection{Result Analysis}
\vspace{-0.1cm}
We compare our extended multi-source weakly supervised saliency detection framework, based on \cite{zeng2019multi}, with existing fully (\textbf{FS}) supervised (pixel-level label), unsupervised (\textbf{US}), and weakly (\textbf{WS}) supervised state-of-the-arts in Tables~\ref{tab:my_label6} and \red{4}. Here, we \emph{additionally} evaluate recently introduced metrics \cite{wu2019stacked} like weighted F-measure $(F_\beta^w)$ and structural similarity measure ($S_\alpha$; $\alpha = 0.5$). Furthermore, we evaluate our sketch-based MS-WSSD framework on the recently introduced more challenging SOC \cite{fan2018salient} dataset.

Following existing weakly supervised methods \cite{zeng2019multi, wang2017learning}, once we train a basic network from weak labels, we use a CRF based post-processing on its predicted output to generate pseudo-ground-truths, which are then used to train a secondary saliency detection network that is finally used for evaluation. Note that our second saliency network directly predicts the saliency map without any post-processing step for faster inference. Qualitative results are shown in Fig.~\ref{fig:qual} using different sources of weak labels, and the Precision-Recall curve on ECSSD and PASCAL-S dataset in Fig.~\ref{fig:iitd}. Overall, we can see that sketches provide a significant edge over category-labels or text-captions and is competitive against recent complex methods like \cite{yu2021structure, liu2021weakly} involving scribbles or bounding-box. In particular, adding class-labels and sketches in the MS-WSSD framework surpasses some popular supervised methods. This success of sketch as a weak saliency label also validates the fine-grained potential of sketches over text/tags in depicting visual concepts.


\vspace{-0.1cm}
\section{Limitations and Future Works}
\vspace{-0.1cm}
 Despite showing how the inherently embedded attentive process in sketches can be translated to saliency maps, our approach, akin to recent sketch literature \cite{xu2020deep, bhunia2020sketch, ghosh2019interactive, zhang2021sketch2model}, involves sketch-photo pairs that are labour-intensive to collect \cite{hu2021towards}. Removing this constraint of a strong pairing deserves immediate future work. Recent works utilising the duality of vector and raster sketches might be a good way forward \cite{sketch2vec}. Furthermore, the study of sketch-induced saliency can also be extended to scene-level, following recent push in the sketch community on scene sketches \cite{zou2018sketchyscene,gao2020sketchycoco, chowdhury2022fs}. This could potentially reveal important insights on scene construction (e.g., what objects are most salient in a scene), and object interactions (e.g., which relationship/activity is most salient), other than just part-level saliency. As our training dataset (Sketchy) contains photo-sketch pairs mainly with a single object, the performance is limited on DUT-OMRON (Table \red{4})  which contains multiple-objects. However, ours is still competitive (or better) compared with most weakly supervised methods \cite{zeng2019multi, wang2017learning, tian2020weakly}. Nevertheless, ours represents the first novel attempt to model saliency from sketches.



\vspace{-0.3cm}
\section{Conclusion}
\vspace{-0.1cm}

We have proposed a framework to generate saliency
maps that utilise sketches as weak labels, circumventing the
need of pixel level annotations. By introducing a sketch
generation network, our model constructs saliency maps
mimicking the human perception. Extensive quantitative
and qualitative results prove our hypothesis – that it is indeed
possible to grasp visual saliency from sketch representation
of images. Our method outperforms several unsupervised
and weakly-supervised state-of-the-arts, and is
comparable to pixel-annotated fully-supervised methods.

{\small
\bibliographystyle{ieee_fullname}
\bibliography{arxiv}
}

\renewcommand{\thefootnote}{\fnsymbol{footnote}}

\onecolumn{
\centering
\title{\Large{\textbf{Supplementary material for \\ Sketch2Saliency: Learning to Detect Salient Objects from Human Drawings}}\vspace{0.3cm}}\\
\author{Ayan Kumar Bhunia\textsuperscript{1} \hspace{.5cm}  Subhadeep Koley\textsuperscript{1,2} \hspace{.5cm} Amandeep Kumar\footnote[1]{Interned with SketchX} \hspace{.5cm} Aneeshan Sain\textsuperscript{1,2} \\ \hspace{.3cm}  Pinaki Nath Chowdhury\textsuperscript{1,2} \hspace{.4cm}
Tao Xiang\textsuperscript{1,2}\hspace{.4cm}  Yi-Zhe Song\textsuperscript{1,2} \\
\textsuperscript{1}SketchX, CVSSP, University of Surrey, United Kingdom.  \\
\textsuperscript{2}iFlyTek-Surrey Joint Research Centre on Artificial Intelligence.}\\
\tt\small \{a.bhunia, s.koley, a.sain, p.chowdhury, t.xiang, y.song\}@surrey.ac.uk

\date{}
}

\maketitle
\thispagestyle{empty} 

\section*{A. Sketch Vector Normalisation}
Firstly, denoting sketch-vectors as a sequence of five-element vectors with \emph{off-set} values over absolute coordinate is common
in sketch/handwriting literature \cite{semi-fgsbir, cao2019ai, song2018learning}. It mainly models the free-flow nature of drawing via GMM \cite{ha2017neural}.
{Regressing to absolute coordinates otherwise, results in mean output \cite{sketchxpixelor} without any instance-specific variation.} Secondly, offset makes sketch invariant to drawing-position in a sketch-canvas (\cref{example_fig}). Keeping rest of the design same, replacing GMM-based loss by {standard $l_{1}$ loss based absolute coordinate regression}, reduces max $F_{\beta}$ value on ECSSD dataset to $0.652$ from $0.781$ (ours), thus validating the need of off-set and GMM-based design. Furthermore, sketch-vector-length varies across samples in a batch, a specific pen-state for end-of-drawing is needed to mask out loss computation from zero-appended tails of sketch-vector.

\vspace{-0.3cm}

\section*{B. Multi-scale 2D Attention Module}
\textit{(i)} $J$ is an intermediate tensor, which is aware of three factors: (a) local and (b) neighbourhood information, of $\mathcal{B}$, and (c) previous state of auto-regressive decoder for sequential modelling. Later, $J \in \mathbb{R}^{\frac{H}{32} \times \frac{W}{32} \times d}$ helps compute the attention-map $\alpha$. \textit{(ii)} $\mathcal{B}(i,j)$ signifies the $1\times1$ convolution applied at every $(i,j)$ spatial position for local-information modelling. \textit{(iii)} The first two terms are tensors of size $\mathbb{R}^{\frac{H}{32} \times \frac{W}{32} \times d}$ and $W_s s_{t-1}$ is a vector of size $\mathbb{R}^{d}$ which is broadcasted (standard PyTorch convention) to the required spatial size for addition. \textit{(iv)} \cref{eqn2} is employed using $1\times1$ convolution with kernel $W_a$, where softmax is applied across the spatial size, $g_t \in \mathbb{R}^{d}$. \textit{(v)} Output of the last three max-pooling layers of VGG-16 are used for multi-scale feature aggregation which have a spatial down-scaling factor of $8, 16, 32$, respectively. \textit{(vi)} During single-scale ablative setup, we only use the output feature-map of the last pooling layer $\mathcal{F}^l$.

\vspace{-0.3cm}

\section*{C. Scribbles vs. Sketch}
Despite taking more time, sketches hold way more structural \textit{and} semantic cues than the much \textit{sparser} and \textit{zero-semantics} scribbles \cite{zhang2020weakly}. Also, temporal aspect of sketches may initiate future works on \emph{relative saliency} of objects at scene level.

\vspace{-0.3cm}

\section*{D. Advantage of Sequential Stroke Modelling}
 Firstly, as free-hand sketches are not edge-aligned  with its paired photo, there is no direct way to post-process the sketch-coordinate to get aligned key-points attending the silhouettes/corners of the object. However, following \cite{zeng2019multi}, we design a baseline as: Apply a spatial attention module on backbone-extracted feature-map followed by global-average pooling and a fully-connected layer to directly predict the $T=100$ (fixed via RDP algorithm) absolute coordinates, assuming that the network will attend salient regions via spatial attention. Here max $F_{\beta}$ on ECSSD dataset falls to $0.692$. Importantly, we can not use off-set based representation here as by definition it relies on sequential modelling explicitly. Therefore, we adopted sequential stroke modelling with `look-back mechanism' via multi-scale 2D-attention for our saliency framework.

\textit{(i)} Moreover, our 1-layer LSTM based auto-regressive network is quite standard for works \cite{semi-fgsbir, bhunia2021joint, cao2019ai, song2018learning}  like image captioning, handwriting/sketch generation, text recognition, and simple to optimise. \textit{(ii)} Removing pen-state prediction hurts accuracy (max $F_{\beta}$: $0.741$) as the model gets confused for large-jumps at stroke-transitions via off-set-based modelling. \textit{(iii)} Furthermore, this temporal aspect of sketch may potentially convey relative saliency (sorted by stroke order) -- an interesting topic for future study.

\vspace{-0.3cm}

\section*{E. Correlating stroke-prediction and saliency-map quality}
We used {photo-to-sketch generation} as an auxiliary task to solve the \textit{saliency detection} problem, where performance of the \textit{latter is crucial} but \textit{not the former}. Moreover we found that, while avg. $F_{\beta}^\text{max}$ for samples whose sketch-generation metric, \emph{log-loss}, (mean of \cref{eq6} in nats \cite{graves2013generating} is lower (better) than $-1100$, comes to $0.824$, the same for samples with \emph{log-loss} higher (worse) than $-1000$ drops to $0.736$, thus justifying the correlation quantitatively.

\end{document}